\documentclass[letterpaper]{article} 
\usepackage{aaai25}  
\usepackage{times}  
\usepackage{helvet}  
\usepackage{courier}  
\usepackage[hyphens]{url}  
\usepackage{graphicx} 
\usepackage{natbib}  
\usepackage{caption} 
\usepackage{algorithm}
\usepackage{algorithmic}
\usepackage{amssymb}
\usepackage{amsthm,amsmath,amssymb}
\usepackage{mathrsfs}
\usepackage{newfloat}
\usepackage{listings}
\DeclareCaptionStyle{ruled}{labelfont=normalfont,labelsep=colon,strut=off} 
\floatstyle{ruled}
\newfloat{listing}{tb}{lst}{}
\floatname{listing}{Listing}
\title{HVIS: A Human-like Vision and Inference System for Human Motion Prediction}

\author{
    Kedi Lyu\textsuperscript{\rm 1,2}, 
    Haipeng Chen\textsuperscript{\rm 1,2}*, 
    Zhenguang Liu\textsuperscript{\rm 3,4}*, 
    Yifang Yin\textsuperscript{\rm 5}, 
    Yukang Lin\textsuperscript{\rm 6}, 
    Yingying Jiao\textsuperscript{\rm 1,2}
}

\affiliations{
	{\normalfont
		\textsuperscript{\rm 1}College of Computer Science and Technology, Jilin University, Changchun, China\\
		\textsuperscript{\rm 2}Key Laboratory of Symbolic Computation and Knowledge Engineering of Ministry of Education, Jilin University.Changchun, China\\
		\textsuperscript{\rm 3}The State Key Laboratory of Blockchain and Data Security, Zhejiang University, Hangzhou, China\\
		\textsuperscript{\rm 4}Hangzhou High-Tech Zone (Binjiang) Institute of Blockchain and Data Security, Hangzhou, China\\
		\textsuperscript{\rm 5}Institute for Infocomm Research (I2R), A*STAR, Singapore\\
		\textsuperscript{\rm 6}Tsinghua Shenzhen International Graduate School, Nanshan District, Shenzhen, China\\
		lvkd19@mails.jlu.edu.cn, 
		chenhp@jlu.edu.cn, 
		liuzhenguang2008 @gmail.com, 
		yin{\_}yifang@i2r.a-star.edu.sg, 
		linyk23@mails.tsinghua.edu.cn, 
		jiaoyy21@mails.jlu.edu.cn	
		}
}

\usepackage{bibentry}

\begin{document}

\maketitle

\begin{abstract}
Grasping the intricacies of human motion, which involve perceiving spatio-temporal dependence and multi-scale effects, is essential for predicting human motion.
While humans inherently possess the requisite skills to navigate this issue, it proves to be markedly more challenging for machines to emulate. 
To bridge the gap, we propose the \textbf{H}uman-like \textbf{V}ision and \textbf{I}nference \textbf{S}ystem (HVIS) for human motion prediction, which is designed to emulate human observation and forecast future movements.
HVIS comprises two components: the \textit{human-like vision encode} (HVE) module and the \textit{human-like motion inference} (HMI) module.
The HVE module mimics and refines the human visual process, incorporating a retina-analog component that captures spatiotemporal information separately to avoid unnecessary crosstalk. Additionally,  a visual cortex-analogy component is designed to hierarchically extract and treat complex motion features, focusing on both global and local features of human poses.
The HMI is employed to simulate the multi-stage learning model of the human brain. The spontaneous learning network simulates the neuronal fracture generation process for the adversarial generation of future motions. Subsequently, the deliberate learning network is optimized for hard-to-train joints to prevent misleading learning.
Experimental results demonstrate that our method achieves new state-of-the-art performance, significantly outperforming existing methods by 19.8\% on Human3.6M, 15.7\% on CMU Mocap, and 11.1\% on G3D. 
\end{abstract}

%

\begin{figure}
\includegraphics[width=\linewidth]{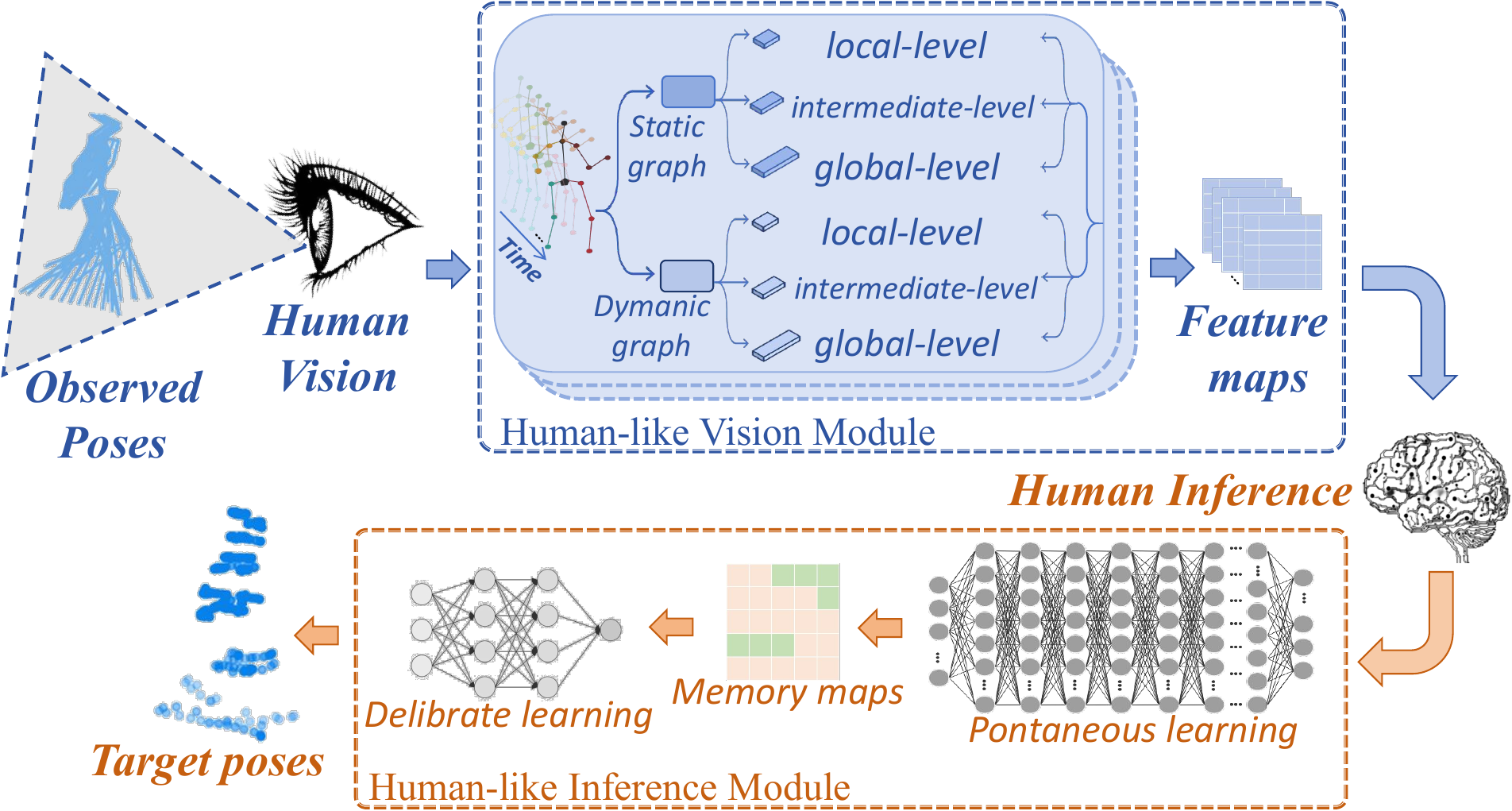}
\caption{Problem illustration. 
}
\label{fig:1}
\end{figure}

\section{Introduction}
Comprehending and predicting human motion constitute an integral aspect of computer vision \cite{Liu1}. Humans exhibit an intuitive understanding of human motion, effortlessly anticipating various ranges of motions and effectively interacting with their physical surroundings (\emph{e.g.}, avoiding crowds on the street). However, this ability is not easily replicated in machines due to the complex relationship between human motion, kinematics, and anatomy. Consequently, building models that enable machines to comprehend and predict human motion is crucial yet challenging. Valid models impel machines to understand and react to human behaviors, which are likewise essential for augmented reality \cite{AR}, animation \cite{animation}, and automatic drive \cite{XunYang01}.

Propelled by the evolution of deep neural networks (DNNS) and the availability of large-scale public motion capture datasets, human motion prediction (HMP) has witnessed extensive investigation and remarkable advancements.
Recurrent neural networks (RNNs) such as Long Short-Term Memory (LSTM) \cite{LSTM}, and Gated Recurrent Unit (GRU) \cite{GRU} have emerged as mainstream solutions for HMP tasks. 
Researchers also leverage anatomical and kinematic constraints to improve performance, incorporating convolutional neural networks (CNNs) \cite{MGCN} due to their ability to capture spatial information. 
Further, generative adversarial networks (GANs) \cite{HPgan} and variational Auto-Encoder (VAE) \cite{vae} are employed to implement non-deterministic predictions. 

Throughout the achievements of HMP, current methodologies are still hindered by three primary challenges:
\quad \textbf{i) The high stochastic nature of human motions:}
Regular motion (\emph{e.g.,} simple harmonic motion, and parabolic motion) can be accurately described by mathematical or physical formulas. Unfortunately, the degrees of freedom (DOFs) of joint movements within space, coupled with unpredictable internal and external stimuli over time, render the description of complex human motions exceedingly challenging using traditional mathematical or physical models. 
Therefore, accurately capturing spatio-temporal dependencies in such a stochastic state is challenging.
\quad \textbf{ii) The high dimensionality of human poses:}
Accurate HMP necessitates precise prediction of the changes in each joint. a pose is consisted of joints $J$ (\text{usually} $J \in [17,56])$. The prior pose sequence $p_p$ has $M$ frames and the future pose sequence $p_f$ has $N$ frames (usually $M,N\in [5,25]$). Accordingly, the dimensionality of a pose sequence $ D=3*J*M ($or $N)$ ($D \in [255,4200]$). 
As a result, such a large amount of data presents an obstacle to understanding the structural characteristics of human bodies.
\quad \textbf{iii) Insufficiency in capturing long-term dependencies:}
Predicting future poses is overly dependent on the RNNS, which leads to errors accumulating during the recurrent process. Additionally, easy-to-train points (e.g., static joints) can be misleading for training. 
Clearly, these issues lead to difficulties in capturing long-term dependencies.

Interestingly, humans are naturally well-equipped to deal with these challenges, such as effective defending an opposing player in a football match. 
This greatly inspires us to emulate human patterns in order to accomplish this task, which can yield substantial advantages for machines to understand and model human motion.
Now the question is \textit{\textbf{how to design a framework to perform human-like HMP task?}}
As a first attempt, we explored human pose regression based on human vision and inference to enhance HMP performance.
\quad \textbf{1) Visual Perception}: 
Given the inherent graph structure of skeleton-based human poses, current methods typically employ spatio-temporal graphs to naturally capture spatio-temporal dependencies. A straightforward extension involves utilizing spatio-temporal maps of all joints in the observed sequence to capture both the structural information of the pose and its temporal dependencies. However, while there is an inherent spatial correlation between neighboring joints in a human pose, the temporal trajectories of individual joints tend to be relatively independent. Thus, the spatial structure of the pose and its temporal dynamics across frames must be captured separately. Additionally, human poses exhibit a distinct hierarchical nature, where single global processing often overlooks local features and significant semantic information.
\quad \textbf{2) Motion Inference}: 
When confronted with complex issues, the human brain typically adopts a multi-stage and multi-strategy learning approach, beginning with an initial comprehensive understanding and progressing to targeted breakthroughs in later stages. In the HMP task, the inference of existing methods often tends to average errors, causing effective features to become obscured within the high-dimensional data. This can sometimes mislead the training process.

To this end, we propose a human-like vision and inference system (HVIS) for HMP, which simulates three aspects of human retinal, visual cortex, and brain learning, respectively, and not only avoids spatiotemporal crosstalk and insufficient local information but also prevents misleading learning. HVIS consists of two modules, \emph{e.g.,} a \textit{human-like vision encode} (HVE) module and a \textit{human-like motion inference} (HMI) module. 
\textbf{In the vision phase}, inspired by the visual pathway \cite{TN} we design the HVE to simulate human hierarchical visual perception to optimize human pose encoding. HVE utilizes a retinotopic analog component (RA) to model the spatial and temporal information of human poses in a discrete manner and a visual cortex analog component (VA) to encode motion information from the RA in a hierarchical manner.
\textbf{In the inference phase}, we designed the HIM to simulate human inference on motion information through two modes of spontaneous and deliberate learning. During spontaneous learning, adversarial learning is utilized to simulate neuronal disconnection and generation. Also considering that the temporal trajectories of each joint tend to be independent, this learning object is joint-level. The deliberate learning network is then targeted mainly at the hard-to-train joints. 
Here we highlight the main contributions as follows:
1) To the best of our knowledge, we are the first to propose a HVIS, which replicates human observation and learning patterns with deep neural networks.
2) We design a human-like vision system that enables human motion modeling along the human visual pathway. Spatio-temporal dependencies as well as global and local information relationships can be adequately captured.
3) We present a novel two-step training strategy for human-like inference, simulating a spontaneous learning process to handle regular prediction processes and a deliberate learning process to enhance hard-to-train joint performance, respectively.
4) Our proposed method achieves state-of-the-art (SOTA) results on three challenging benchmark datasets, namely H3.6M, CMU, and G3D.

\section{Related Work}
\textbf{GCN-based Methods.}\quad
The skeletal representation of humans, characterized by joints and their adjacent links, can be effectively modeled as a graph. This correlation has been extensively explored by researchers \cite{LTD,HRI,yao2024swift}. Graph Convolutional Networks (GCNs), specifically designed for graph data, offer a robust method for feature extraction in this context.
In \cite{DMGNN,SKEL}, a multi-scale graph was introduced that adeptly captures features across multiple scales, enabling the accurate prediction of future human motions. Concurrently, \cite{DBLP:conf/iccv/LiuS0SCH021} developed a semi-constrained graph that explicitly encodes skeletal connections and integrates prior knowledge. Building on this, they proposed a multi-scale spatial-temporal graph \cite{MSGNN} to achieve comprehensive modeling.
Differently, our work simulate the human visual system. We design a unique GCN structure that handles dynamic and static information separately, thereby eliminating the disruption of irrelevant data. Moreover, we encode the data hierarchically to obtain richer localized human information.

\textbf{TCN-based methods.} \quad
RNNs are inherently prone to error accumulation \cite{QN,SGRU,MM23_deepfake}. Conversely, Temporal Recurrent Networks (TCNs) mitigate common issues such as gradient explosion or vanishing gradients, as their back-propagation paths diverge from the temporal sequence direction. These strengths have led to the increasing application of TCNs in HMP.  
For instance, in \cite{MoP}, TCNs are employed to decode the dynamics of sub-motions and the spatial correlations within the entire motion sequence to predict future movements. Subsequently, \cite{TCGAN} propounds a residual TCN that is characterized by a minimalist design yet delivers high efficiency.
Diverging from conventional methods, we have developed a TCN-based model. This network harmoniously fuses the strengths of both RNNs and TCNs, creating a recurrent temporal convolutional network adept at predicting future movements. 

\textbf{GAN-based Methods.} \quad
GANs have emerged as a pivotal tool in addressing HMP challenges, not only for constructing novel networks but also for advancing learning algorithms. Recent advancements include specialized adaptations such as HP-GAN, which modifies the improved WGAN-GP for probabilistic HMP \cite{HPgan}. Additionally, Bi-GANs \cite{Bigan} and GAN-poser \cite{Gan-poser} introduce bi-directional structures to enhance prediction accuracy. An AMGAN \cite{AMGAN} targets kinematic chains.
Adversarial strategies in motion analysis also be leveraged for network training \cite{ARNET, AGED, yao2024swift}, demonstrating their effectiveness in various contexts. Notably, SGRU \cite{SGRU} employs GANs to simulate path integral, offering a distinct perspective on motion modeling.
In our investigation, we contribute a theoretical demonstration that dimension reduction significantly enhances the convergence of WGAN. Moreover, we utilize GANs for motion context modeling.

\begin{figure*}[t]
\begin{center}
\includegraphics[width=.9\linewidth]{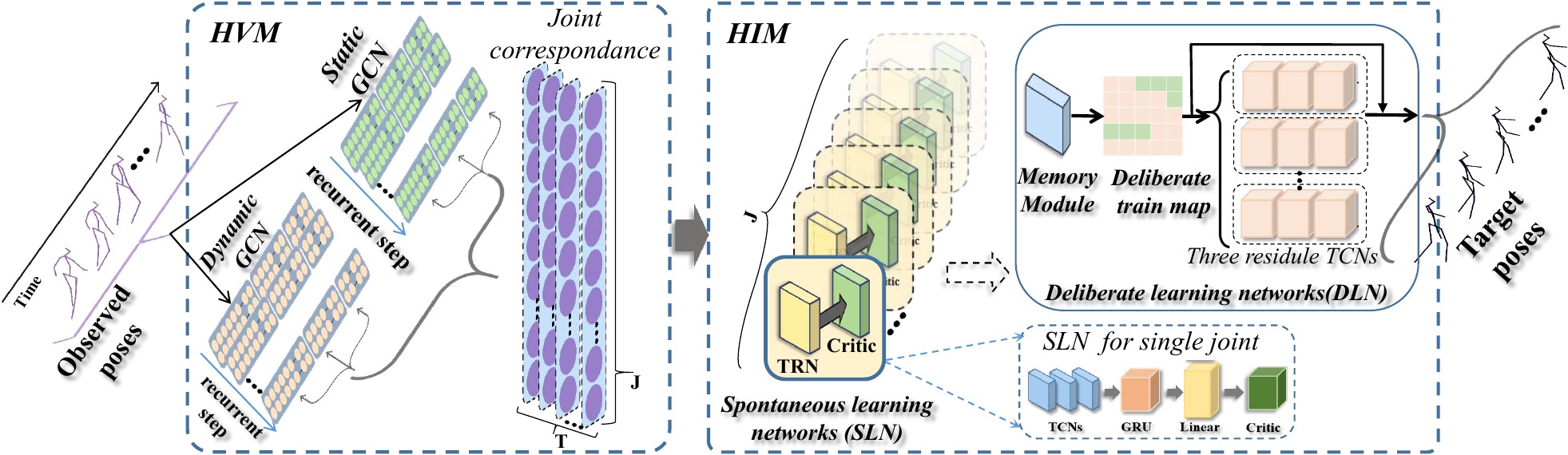}
\caption{HVIS including two components: Human-like vision module (HVM) and Human-like inference module (HIM).}
\label{fig:2}
\end{center}
\end{figure*}

\section{Our Approach}

\textbf{Problem Formulation.} \quad
Given an observed human pose sequence $( P_1, P_2,..., P_O )$ with $O$ human poses. The objective of HMP task is to predict the future human motion sequence $( P_{O+1}, P_{O+2},..., P_{O+F} )$ in the next $F$ frames. 

\textbf{Method Overview.} \quad
As a first attempt, HVIS predicts 3D human motion from an observed human pose sequence. The key idea is to establish a human-like motion prediction framework to aid in understanding and modeling human motions. 
The pipeline of our proposed approach is illustrated in Fig.\ref{fig:2} with two meticulously designed components, namely \textit{human-like vision module} (HVM) and \textit{human-like inference module} (HIM). Specifically, the former is proposed to encode human motion, capturing spatiotemporal dependencies in a separate manner that mimics retinal properties, and emulating the optic cortex in its hierarchical processing of information. The latter is designed to simulate the learning patterns of the human brain and model the context from HVM by spontaneous and deliberate network learning.
In what follows, we will elaborate on the technical details of the two components, respectively.

\subsection{Human-like Vision Module}
Traditional methods represent parameterized 3D human pose sequences as kinematic spatio-temporal graphs \cite{ST-gcn,MSGNN,MGCN}, leveraging the inherent structural properties of skeleton-based human poses. Through a detailed analysis of human pose data and its encoding, we identify two key issues:
\quad i) Despite the inherent spatial correlation between neighboring joints in a human pose, the temporal trajectories of each joint are often quite independent. This necessitates a separate capture of the spatial structure of the pose and its temporal dynamics across frames.
\quad ii) In high-dimensional data, localized information is often obscured, so it becomes difficult for the network to fully extract this useful hierarchical information.

Given an observed human pose sequence $\Gamma=( P_1, P_2,..., P_T )$, consisting of $T$ poses $P_t^N (t \in [1,T] )$, where $N$ denotes the number of joints, we encode the motion history tensor into a graph. This graph is constructed with an intent to model the intricately interwoven interactions among all body joints across every captured frame. 
The encoding graph is defined as $G=(V,E)$, wherein $V$ represents nodes - the total $J=N*T$ body joints across all observed time frames. The connections or edges are symbolized by the spatio-temporal adjacency matrix $A^{st} \in R^{J^2}$, delineating the interactions of all joints at all times. 

Ordinarily, the spatio-temporal dependencies of joints can be encoded efficiently through a Graph Convolutional Network (GCN). The input to a graph convolutional layer $L$ is the tensor $T_{(L)}\in R^{C_L\times J}$, encoding the observed $J$ joints, with $C_L$ being the input dimensionality of the hidden representation $T_{(L)}$. A graph convolutional layer $L$ then produces the output $T_{(L+1)}\in R^{C_{L+1}\times J}$, which can be obtained from the input $T_{(L)}$ of the current layer, by multiplying it with the corresponding weights $W_{L}$ and the shape space adjacency matrix $A^{st}_{L}$ and through the activation function $\mathcal{G}$.

Our approach aims to simulate the human visual pathway \cite{TN} for encoding human poses. The human visual system processes information hierarchically, beginning at the retina and moving through the primary visual cortex to higher-order visual cortex areas, with each stage further refining and abstracting visual input. We introduce a \textbf{retinotopic analog component} (RA), which processes pose data structured graphically in a discrete manner, leveraging the principles of optic cone and sensory cell. This design mitigates spatio-temporal crosstalk and effectively captures spatio-temporal dependencies. A \textbf{visual cortex analog component} (VA) hierarchically processes data from the RA to prevent critical information from being overwhelmed and fully access global and local information.

\textbf{Retinotopic Analog Component (RA).}\quad
RA targets to mimic the synergy of optic cone cells and optic sensory cells by decomposing the space-time adjacency matrix into the product of static adjacency matrices $A^s$ and dynamic adjacency matrices $A^d$. Consequently, a RA graph convolutional layer $L$, its output $T_{\textit{L}}$ is defined as:

\begin{equation}
\begin{aligned}
T_{(L+1)}=\mathcal{G}(D_L^{-1/2} A_L^s A_L^d D_L^{-1/2} T_L W_L)
\end{aligned}
\label{Eq:system}
\end{equation}
The adjacency matrix $A^s$ accounts for the static interactions, while $A^d$ handles the dynamic relations. During graph convolution, the distinct implications of static and dynamic interactions are taken into account. 

\textbf{Visual Cortex Analog Component (VA).}\quad
The purpose of VA is to prevent information flooding by hierarchically encoding the human graph structural information and to fully capture global and local information. It simulates the work pattern of the human visual cortex, where different levels of the cortex also process information of different scales and complexity. Directly, it is to consider the spatial and temporal information on different scales so as to better capture the complex dynamics of human poses. We need to construct multiple spatial and temporal adjacency matrices at different scales. Suppose we have $M$ scales, and the corresponding spatial and temporal adjacency matrices for each scale are $\{A_s^{(m)}\}_{m=1}^M$ and $\{A_t^{(m)}\}_{m=1}^M$. Consequently, a VA graph convolutional layer $L$, its output $T_{\textit{L}}$ is defined as:

\begin{equation}
\begin{aligned}
T_{(L+1)}=\mathcal{G}(\sum_{m=1}^M{A_{(s)L}^{(m)} A_{(t)L}^{(m)} T_L^{(m)} W_L^{(m)}})
\end{aligned}
\label{Eq:system}
\end{equation}

Within our approach, the number of scales, $M=3$, are processed separately for different scales of information. The primary visual cortex corresponds to the processing of joint-scale information, the intermediate visual cortex corresponds to the processing of kinematic chain-scale (trunk and limbs) information, and the advanced visual cortex corresponds to the processing of overall pose information.

\subsection{Human-like Inference Module} 
So far, we have explored the utilization of HVM with spatio-temporal information separation and hierarchical representation to comprehensively capture spatio-temporal dependencies and human poses, from local to global information, for advanced human pose coding. However, the challenges of high dimensionality in pose sequences and the misleading effects of easily trainable joints on network learning remain significant issues. In response, we propose the Human-like Inference Module (HIM) to simulate the learning mode of the human brain and predict future motion through spontaneous and intentional learning. 
HIM includes two parts: a \textbf{spontaneous learning network} (SLN), and a \textbf{deliberate learning network} (DLN). 
SLN simulates the form of neuronal fracture synthesis in human brain learning to design joint-level adversarial networks to generate future human motions to alleviate the distress caused by the high dimensionality of pose sequences. 
DLN, as the name implies, prevents the emergence of misleading training by actively filtering the easy and difficult training points for targeted training.

\textbf{Spontaneous Learning Network (SLN).} \quad
The SLN is to effectively manage the high dimension inherent in human poses and mitigate the error accumulation typically associated with traditional RNN-based modeling. For these, two sub-modules are designed i.e. a \textit{temporal recurrent network} (TRN) and a \textit{critic}. Note that adversarial networks are not solely inspired by the workings of the human brain. Our approach leverages a theoretical insight related to GANs that helps to address the dimensionality issue. The specific theoretical findings are shown in the following section.

\textit{Curse of Dimensionality.} \quad
In \cite{Dudley1969TheSO}, it shows that the the rate of convergence of $v_n$ to $v$ in the Wasserstein-1 metric is $n^{-1/d}$. 
\begin{equation}
\begin{aligned}
\mathbb{E}[W_1(v,\hat{\textit{v}}_n)] \asymp n^{-1/d}.
\end{aligned}
\end{equation}
\cite{2019Sharp} extends these results to the Wasserstein-p metric, demonstrating that the rate of convergence slows exponentially as data dimensionality increases, highlighting the curse of dimensionality.
Meanwhile, we account for the independence of human joints in temporal sequences.
Based on these, SLN is deployed to each joint of the human pose with a \textit{TRN} as the generator and a \textit{critic}.  

\textit{TRN.}\quad 
Based on the above, TRN is designed with a temporal information Unit (TIU) and a latent temporal features Unit (LTF). 
In detail, TIU includes three parts. 
First, the TCN blocks serve as the foundational element due to their robustness against error accumulation. Notably, the more widely adopted Transformer framework was deliberately not chosen for this task, owing to its inconsistent performance on HMP task. The discussion on this decision is detailed in the Experimental section.
Then, a residual learning framework is employed among the TCN blocks to improve the training efficiency.
Finally, dilated convolution is utilized to capture sparser temporal information.
Following these, the TIU can target the \textit{joint-level} information $S^i$ from the HVM and generate latent temporal features $L_{S^i}$ to be fed to the LTF.
In the LTF, we opt for the GRU over the TCN for processing latent information $L_{S^i}$. This decision is informed by empirical evidence demonstrating that GRUs are more effective in capturing and processing latent temporal information. Specific experimental proofs are available in our open-source library.
This procedure can be expressed as:

\begin{equation}
\begin{aligned}
S^i \xrightarrow[{}]{\textit{3 residual TCN blocks}}L_{S^i}\xrightarrow[{}]{LTF(\cdot)}S^{i+T}.
\end{aligned}
\label{eq:lp}
\end{equation}

\textit{Critic.}\quad 
A single-layer fully connected feed-forward network is utilized as the critic. First, it measures the similitude of the distributions between the generated pose and the ground truth. Second, the critic assesses whether generations are natural and smooth.

\textbf{Deliberate Learning Network (DLN).}\quad
Our training methodology aims to minimize the mean error; however, the network struggles to differentiate between easy-to-train points (e.g., static joints) and difficult-to-train points (e.g., joints with irregular motions). This often leads to suboptimal training performance. To address this issue, we adopt a strategy of deliberate training, which involves explicitly memorizing the hard-to-train joints and concentrating the learning process on these challenging joints.
A \textit{deliberate learning Network} (DLN) is proposed, which includes a \textit{memory component} (MC) and a \textit{deliberate training component} (DTC).

\textit{MC.}\quad 
In the MC, a trained TRN is employed as the memory component $\Phi (\cdot)$, where all the joints $\{S^i\}i_{i=1}^N$ are fed, $N$ is the number of joints, as shown in Eq \ref{eq:mm1}.  
\begin{equation}
\begin{aligned}
U_{\textit{ep}} = \Phi (\{S^i\}_{i=1}^N).
\end{aligned}
\label{eq:mm1}
\end{equation}
Then, memory component $U_\textit{ep}$ are ranked quantitatively by the rank function $\mathcal{R}(\cdot) $ to get a deliberate train map $\Psi$. 
\begin{equation}
\begin{aligned}
\Psi = \mathcal{R}(U_{\textit{ep}}).
\end{aligned}
\end{equation}
The $\Psi$ can guide the network to get corresponding joint $\{S_M^i\}_{i=1}^m$, $m$ is the number of hard-to-train joints.
\begin{equation}
\begin{aligned}
\{S_M^i\}_{i=1}^m = \Phi (\{S^i\}_{i=1}^N, \Psi).
\end{aligned}
\end{equation}
Finally, these $S_M^i$ are inputted into the DTC.

\textit{DTC.}\quad
The objective of the DTC is to \textbf{only} train the $\{S_M^i\}_{i=1}^m$ deliberately and generate the target $ \{\hat{S}_M^i\}_{i=1}^m$. In order to handle the temporal information efficiently, we design the DTM consisting of multiple TCN blocks.

\subsection{Loss Function}
Until now, we have introduced our framework for predicting human motions. Now, we focus on the loss function for training and optimizing our model.
The objective is to minimize the error between generation and target joints. There are mainly three loss functions utilized in the training process: \textit{generator loss}, \textit{critic loss}, and \textit{deliberate train loss}. 

\textbf{Generator Loss.} \quad
It is similar to WGAN adversary loss, with an added joint error loss $L_{S_i}$ shown as follows:
\begin{equation}
\begin{aligned}
L_\textit{G} = L_{\textit{wg}} + L_{S_i}
\end{aligned}
\end{equation}
where $L_{\textit{wg}}$ is the generator loss in the WGAN defined as:
\begin{equation}
\begin{aligned}
L_{\textit{wg}} = -\textit{D}[(\hat{S}^{t+1},...,\hat{S}^{t+T} )|(S^1,...,S^t)]
\end{aligned}
\end{equation}
where $S^{[1:t]}$ is the observed joints and
$\hat{S}^{[t+1:t+T]}$ is the generated joints. And the joint loss $L_{j}$ is defined as: 

\begin{equation}
\begin{aligned}
L_{j} = \frac{1}{T}\sum_1^T|S^{t+i}-\hat{S}^{t+i}|^2
\end{aligned}
\end{equation}

\textbf{Critic Loss.} \quad
The Critic attempts to award higher scores to the real future joints and lower scores to the generated joints. Follow this, the critic loss is defined as: 
\begin{equation}
\begin{aligned}
L_{critic} = \textbf{E}\textit{D}(S^{t+1:t+T})-\textbf{E}\textit{D}(\hat{S}^{t+1:t+T})
\end{aligned}
\end{equation}
where $S$ is the target joints and $\hat{S}$ is the generated joints.

\textbf{Deliberate Train Loss.}\quad
It is responsible for the individual training of the memorised joints.
\begin{equation}
\begin{aligned}
L_{\textit{D-train}} = \frac{1}{T}\sum_1^T|S_M^{t+i}-\hat{S}^{t+i}_\textit{M}|^2
\end{aligned}
\end{equation}
where $\hat{S}^{t+i}_\textit{M}$ represents the memorised joints.


\begin{table*}[t]
	\centering
	\resizebox{\textwidth}{!}
	{
		\renewcommand\scriptsize{7pt}
            
		\begin{tabular}{ccccccccccccccccccccl}
            \hline
            Time (ms) & 80 & 160 & 320 & 400 & 1,000 & 80 & 160 & 320 & 400 & 1,000 & 80 & 160 & 320 & 400 & 1,000 & 80 & 160 & 320 & 400 & 1,000 \\ 
			\hline
			&\multicolumn{5}{c}{Directions}  & \multicolumn{5}{c}{Greeting}    & \multicolumn{5}{c}{Phoning}         & \multicolumn{5}{c}{Posing}      \\ 
			res-GRU 
			&36.4 &56.6 & 80.3& 98.1&126.3 
			&36.8 &73.3 & 138.2& 155.6& 189.5
			&24.3 &42.3 & 72.6& 82.3& 124.2
			&26.7 &52.4 & 129.5& 159.4& 181.7 \\ 
			
			HPGAN
			& 80.9& 101.3& 148.6& 168.8 &  234.6 
			&81.5 & 118.8 & 178.4& 200.1 & 258.6
			&78.8 & 100.3 & 152.7 & 179.0 & 244.2
			&75.5 & 107.4 & 168.3 & 178.0 & 250.1  \\ 
			
			BiGAN 
			& 22.0 & 37.5 & 58.9& 72.0 & 114.7
			& 24.6 & 45.8 & 89.9 & 103.0 & 148.1
			& 17.0 & 29.7 & 54.1 & 62.1 & 112.0 
			& 16.8 & 35.0 & 86.4 &105.6 & 187.0\\
			
			HMR 
			& 23.3 & 25.0& 47.2 & 61.5 &  116.9
			&12.9 & 31.9 & 55.6& 82.5 & 123.2
			&12.5 & 21.3 & 39.3 & 58.6 & 112.8
			&13.6 & 23.5 & 62.5 & 114.1 & 143.6 \\ 
			
			LTD 
			&9.2& 20.6 & 46.9 & 58.8 & 105.8  
			&16.7& 33.9& 67.5& 81.6& 140.2
			&10.2&20.2& 40.9&50.9& 105.1
			&12.5 &27.5& 62.5& 79.6& 171.7 \\ 
			
			DMGNN 
			& 12.3& 23.8& 46.2& 55.5&90.3 
			& 14.0& 29.8& 74.0& 89.1& 140.2
			& 10.2& 14.0& 32.8& 40.0& 104.1
			&9.2 & 23.5& 65.0& 82.8& 170.2 \\
			
		    HRI 
			& 7.4 & 18.4 & 44.5& 56.5 & 106.5
			& 13.7& 30.1 & 63.8 & 78.1 & 138.8
			& 8.6 & 18.3  & 39.0 & 49.2 & 105.0
			& 10.2 & 24.4 & 58.5 &75.8& 178.2\\	
			
			MSR-GCN 
			& 8.6& 19.7& 43.3& 53.8& -
			& 16.5& 37.0& 77.3& 93.4& -
			& 10.1& 20.7& 41.5& 51.3& -
			&12.8& 29.4& 67.0&85.0 &- \\

                SPGSN
                & 7.4 & 16.4 & 39.6& 50.1 & 97.2
			& 14.6 &32.6 &70.6 &86.4 & 143.2
			& 8.7 &18.3 &38.7 &48.5 & 102.5
			& 10.7 &25.3 &59.9 &76.5 & 165.4  \\

                FDU
                & 6.6 &16.4 &39.6 &50.1 & 97.2
			& 13.0 &30.7 &63.1 &78.24 & 141.8
			& 7.8 &17.2 &37.5 &47.3 & 96.7
			& \textbf{7.5} &19.3 &47.1 &\textbf{62.0} & 149.5 \\
			
			Ours
			& \textbf{6.3 }& \textbf{10.7} & \textbf{17.1 }& \textbf{29.8} & \textbf{59.8}
			& \textbf{9.2} &\textbf{ 25.3} & \textbf{40.1} & \textbf{63.2} & \textbf{102.3}
			& \textbf{6.7} & \textbf{12.8 }& \textbf{30.9} & \textbf{42.5} & \textbf{78.7}
			& 9.6 & \textbf{18.8}& \textbf{45.2} &70.0 & \textbf{108.2}\\ 
			\hline
			
			& \multicolumn{5}{c}{Waiting} & \multicolumn{5}{c}{Eating}
			& \multicolumn{5}{c}{Smoking} & \multicolumn{5}{c}{Discussion}  \\ 
						
			res-GRU 
			&20.5 & 39.8& 78.2 & 90.3& 120.1 
			&17.5 & 34.3& 71.1 & 87.5& 117.6
			&22.4 &39.9 & 80.2 & 92.5& 119.2
			&25.8 &43.4& 83.5& 95.8 & 129.1 \\ 
			
			HPGAN 
			&70.1& 89.6& 98.2 & 121.0 &  145.2
			&64.1 & 78.4 & 99.9& 113.7 & 136.2
			&67.2 & 88.6 & 100.1 & 123.9 & 140.4
			&71.4 & 91.3 & 105.2 & 129.7 & 150.4  \\ 
			
			BiGAN 
			& 17.5 & 31.3 & 53.9& 61.4 & 128.5 
			& 13.6 & 26.1 & 51.4 & 63.1 & 84.1
			& 11.0 & 21.0 & 33.1 & 38.2 & 88.1
			& 19.2& 39.0 & 67.7 &75.3 & 122.5\\

			HMR 
			& 17.2 & 31.4 & 53.5& 61.1 & 99.0 
			& 13.2 & 26.0 & 51.1 & 62.6 & 74.0
			& 10.3 & 20.5 & 33.0 & 37.2 & 69.1
			& 19.0& 38.8  & 67.3 & 75.0 & 121.5\\
			
			LTD 
			& 10.5& 21.6& 45.9& 57.1&106.9
			&7.7 & 15.8& 30.54& 37.6& 74.1
			&8.4&16.8& 32.5&39.5& 73.6
			&12.2 &25.8& 53.9& 66.7& 118.6\\ 
			
			DMGNN 
			& 12.2& 24.1& 60.0& 77.5& 128.0
			&11.0& 21.4& 36.1& 43.9 & 57.0
			&9.0& 17.6& 25.1& 40.3 & -
			&17.3 &34.8& 61.0& 70.0 & - \\ 
			
			HRI 
			&8.7& 19.2& 43.4& 54.9&108.2
			&8.7& 18.7& 39.5& 47.1 & 57.0
			&7.0& 14.9& 29.9& 36.4 & 69.5
			&10.2 &23.4& 52.1& 65.4 & 119.8 \\
			
			MSR-GCN 
			&10.7& 23.1& 48.3& 59.2& -
			&8.4& 17.1& 33.0& 40.0 & -
			&8.0& 16.3& 31.3& 38.2 & -
			&12.0 &26.8& 57.1& 70.0 & - \\

                SPGSN
                & 9.2 &19.8 &43.1 &54.1 & 103.6
			& 7.1 &14.9 &30.5 &37.9 & 73.4
			& 6.7 &13.8 &28.0 &34.6 &  68.6
			& 10.4 &23.8 &53.6 &67.1 &  118.6  \\

                FDU
                & 8.2 &18.4 &41.3 &52.1 & 101.2
			& 6.3 &13.7 &29.1 &36.3 & 71.1
			& \textbf{5.1} & \textbf{9.1} &21.3 &29.9 & \textbf{59.3}
			& 7.4 &17.1 &42.9 &50.4 & 92.3 \\
			
			Ours 
			& \textbf{7.3} & \textbf{15.8} & \textbf{38.1} & \textbf{49.5} & \textbf{92.6}
			& \textbf{6.0} & \textbf{12.7}& \textbf{21.2} & \textbf{27.1} &\textbf{57.3}
			& 5.9 & 10.1  & \textbf{20.2} & \textbf{27.5} & 61.2
			& \textbf{7.1} & \textbf{16.8}& \textbf{31.2} &\textbf{46.0} & \textbf{87.4}\\
			
			\hline
			
			& \multicolumn{5}{c}{Purchases}&\multicolumn{5}{c}{Sitting} & \multicolumn{5}{c}{Sittingdown}&\multicolumn{5}{c}{Takingphoto} \\ 
			
			res-GRU 
			& 38.5& 70.1& 101.0& 102.3&131.2 
			&34.1 & 53.2& 110.4& 115.0& 150.1
			&28.6 &55.2& 85.6&115.8 & 180.0
			&23.1 &47.0& 92.3& 110.1& 149.2 \\ 
			
			HPGAN 
			& 42.4& 88.9& 95.0 & 120.2 &  170.2
			&36.3 & 60.0 & 120.0& 123.1 & 168.2
			&39.9& 65.9 & 92.1 & 130.0 & 200.2
			&38.0 & 49.3 & 79.9 & 83.8& 160.4  \\ 
			
			BiGAN 
			& 29.0 & 54.1 & 82.2 & 92.4& 139.0
			& 19.9 & 41.0 & 76.3 & 88.2 & 120.5 
			& 17.0 & 34.8  & 66.5& 76.9 & 152.0
			& 14.2 & 27.1 & 53.5 &66.1 & 128.0\\

			HMR 
			& 15.3 & 30.6& 64.7 & 73.9 &  122.7 
			&12.6 & 25.6 & 44.7& 60.7 & 118.4 
			&9.6 & 18.6 & 41.1 & 57.7 & 148.3
			&7.9 & 19.0 & 31.5 & 57.3 & 108.5\\ 
			
			LTD 
			&15.5& 32.3& 64.9& 78.1&135.9
			&10.7& 24.6& 50.6& 62.0& 115.7
			&17.0&33.4& 61.6&74.4& 144.1
			&9.9 &20.5& 43.8& 55.2& 120.2\\ 
			
			DMGNN 
			& 21.4& 38.7& 75.7& 92.7&- 
			&11.9& 25.1& 44.6& 50.2& -
			&15.0&32.9&77.1&93.0& -
			&13.6 &29.0& 46.0& 58.8& - \\ 
			
			HRI 
			& 13.0& 29.2& 60.4& 73.9& 134.2
			&9.3& 20.1& 44.3& 56.0 & 115.9
			&14.9& 30.7& 59.1& 72.0 & 143.6
			&8.3 &18.4& 40.7& 51.5 & 115.9 \\
			
			MSR-GCN
			& 14.8& 32.4& 66.1& 79.6& -
			&10.3& 22.0& 46.3& 57.8 & -
			&16.1& 31.6& 62.5& 76.8 & -
			&9.9 &21.0& 44.6& 56.3 & - \\

                SPGSN
                & 12.8 &28.6 &61.0 &74.4 &133.9
			& 9.3 &19.4 &42.3 &53.6 &116.2
			& 14.2 &27.7 &56.8 &70.7 & 149.9
			& 8.7 &18.9 &41.5 &52.7 &118.2 \\

                FDU
                & 11.8 &27.2 &56.4 &63.9 &130.7
			& 8.7 &18.9 &42.1 &53.2 &114.5
			& 13.9 &25.6 &54.2 &67.2 & 145.3
			& 8.1 &18.0 &39.2 &50.6 & 116.1\\
			
			Ours
			&\textbf{11.0}& \textbf{26.8 }& \textbf{50.2} & \textbf{60.5} & \textbf{105.8}
			& \textbf{8.2} & \textbf{18.1} &\textbf{37.2} & \textbf{48.9} & \textbf{108.4} 
			& \textbf{8.9} & \textbf{17.8}  & \textbf{38.2} & \textbf{55.6} & \textbf{99.8}
			& \textbf{7.8}& \textbf{13.5} & \textbf{27.2} &\textbf{43.1} & \textbf{94.1}\\ 
			\hline

	\end{tabular}}
 \caption{Performance evaluation (in MPJPE) on the H3.6m dataset. The best results are highlighted in bold.}
 \label{tab:tab01}
\end{table*}

\begin{table*}[t]
	
	\centering
	\resizebox{\textwidth}{!}{
		\renewcommand\scriptsize{7.0t}
		\begin{tabular}{ccccccccccccccccccccc}
            \hline
             Time (ms) & 80 & 160 & 320 & 400 & 1,000 & 80 & 160 & 320 & 400 & 1,000 & 80 & 160 & 320 & 400 & 1,000 & 80 & 160 & 320 & 400 & 1,000 \\ 
			\hline
			& \multicolumn{5}{c}{Basketball}  & \multicolumn{5}{c}{Basketball Signal}    & \multicolumn{5}{c}{Directing Traffic}         & \multicolumn{5}{c}{Jumping}      \\ 
            
			
			res-GRU  & 18.5& 33.9& 48.1& 49.0& 106.3 &12.9& 23.8& 40.2& 60.1& 77.5& 15.6& 30.1& 55.2 &66.1& 127.1& 36.1& 68.7& 125.0& 140.0& 192.6\\ 
			
			BiGAN  & 16.5& 30.5& 47.2& 48.8& 91.5 &8.7& 16.3& 30.1& 37.8& 76.6& 10.6& 20.3& 38.7&49.0& 113.3& 22.4& 44.3& 87.3& 105.1& 156.3\\ 
			
			LTD  &14.3& 25.5& 48.4& 62.6& 109.0 &3.5& 6.7& 12.0& 15.8 &54.4 &7.4& 15.5& 31.9& 42.5 &151.9 &16.9& 34.4& 76.3& 98.6& 164.4 \\ 
			
			MSR-GCN  & 13.1& 22.1& 37.2& 55.8& 97.7 &3.4& 6.2& 11.2& 13.8& 47.3& 6.8& 16.3& 66.3 &78.8& 129.7& 11.0& 24.5& 65.7& 90.3& 189.1\\ 
			
			Ours & \textbf{10.5} & \textbf{19.3}& \textbf{35.5} & \textbf{46.8} & \textbf{89.3} 
			& \textbf{2.5} & \textbf{6.0 }& \textbf{10.9 }& \textbf{12.8} & \textbf{44.5}
			& \textbf{4.9}  & \textbf{9.8 } & \textbf{21.9} & \textbf{29.7} & \textbf{98.5}
			& \textbf{10.4}& \textbf{23.5} & \textbf{60.1} & \textbf{85.6} & \textbf{148.2}\\ 
			\hline
			
			& \multicolumn{5}{c}{Running}  & \multicolumn{5}{c}{Soccer}     & \multicolumn{5}{c}{Walking}    & \multicolumn{5}{c}{Washing Window} \\ 
            
			
			res-GRU  & 17.4& 20.0& 27.3& 36.7& 50.2 &
			20.3 & 39.5& 71.3& 84.0& 129.6& 8.2& 13.7& 21.9 &24.5& 32.2 & 8.4& 15.8& 29.3& 35.4& 61.1\\ 
			
			BiGAN   & 14.3& 16.3& 18.0& 20.2& 27.5 &12.1& 21.8& 41.9& 52.9& 94.6& 7.6& 12.5& 23.0&27.5& 49.8& 8.2& 15.9& 32.1& 39.9& 58.9\\ 
			
			LTD    &25.5& 36.7& 39.3& 39.9& 58.2 &11.3& 21.5& 44.2& 55.8 &117.5&7.7& 11.8& 19.4& 23.1 & 40.2 &5.9& 11.9& 30.3& 40.0& 79.3 \\ 
			
			MSR-GCN   & 15.2& 19.7& 23.3& 35.8& 47.4 &10.3& 21.1& 42.7& 50.9& 91.4& 7.1& 10.4& 17.8 &20.7& 37.5& 5.8& 12.3& 27.8& 38.2& 56.6\\ 
			
			Ours & \textbf{11.1} & \textbf{14.3} & \textbf{17.2} & \textbf{18.8} & \textbf{25.4}
			& \textbf{7.9}& \textbf{15.1}& \textbf{30.5} & \textbf{41.2} & \textbf{85.3} 
			& \textbf{5.8}& \textbf{8.5} & \textbf{15.1} & \textbf{17.2} & \textbf{30.1} 
			& \textbf{4.5}& \textbf{9.2} & \textbf{26.1} & \textbf{32.2} & \textbf{55.1}\\ \hline
			
	\end{tabular}}
 \caption{Performance evaluation (in MPJPE) on CMU MoCap dataset. The best results are highlighted in bold.}
 \label{tab:tab02}
\end{table*}

\begin{figure*}[h]
	\centering
	\includegraphics[width=.7\textwidth]{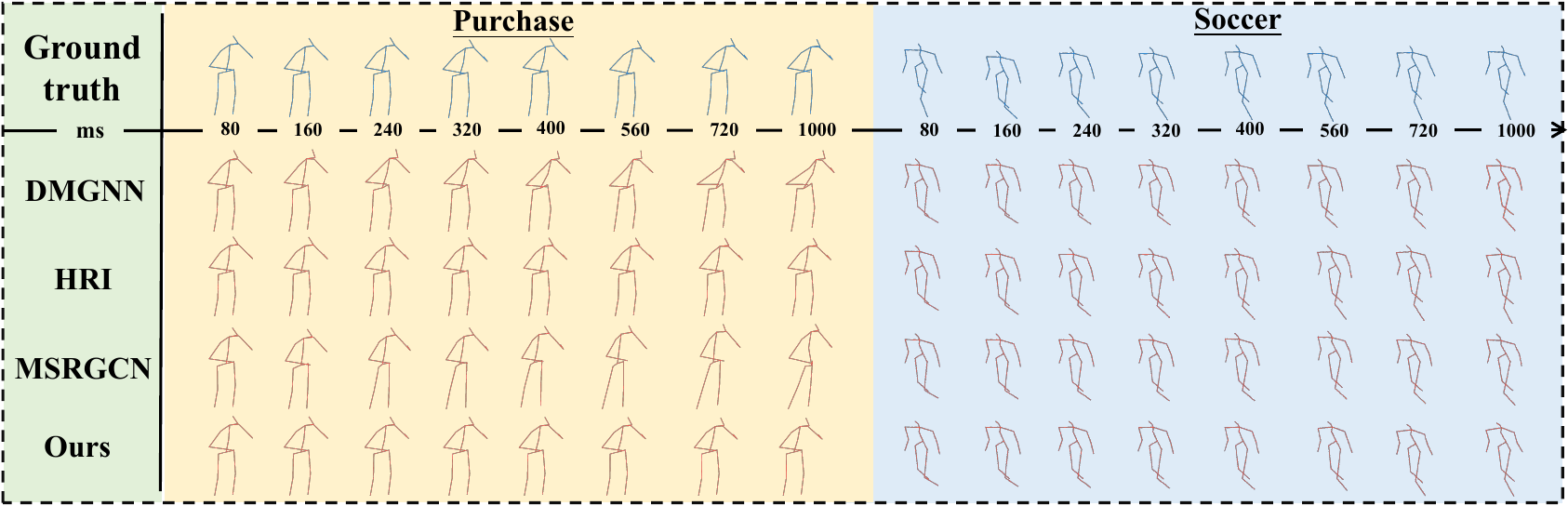}
	\caption{Visual comparisons on H3.6M dataset (Purchase) and CMU dataset (Soccer). The blue poses are the ground truth.}
	\label{fig:3}
\end{figure*}

\section{Experiments}

\subsection{Datasets and Experimental Settings}

\textbf{Datasets.}\quad To testify the effectiveness and robustness of our proposed model, three large benchmark datasets Human 3.6 Million (H3.6M), CMU MoCap (CMU), and G3D are engaged.
\textbf{H3.6M} public dataset records a total of 3.6 million human motion data involving 15 different actions. Experimentally, these poses are divided into 7 subjects and removed duplicate points of the human pose. A down-sampling is set to 25 FPS.
\textbf{CMU} dataset has 144 different subjects. In general, these samples are split into a training set and a test set in the experiment. The sequences are also 25 FPS.
\textbf{G3D} is a gaming dataset from Microsoft Kinect devices and Windows SDK. Totally, 210 samples and 10 subjects perform 20 gaming action. The frame rate is 30 FPS.

\textbf{Experimental Settings and Evaluation Metrics.}\quad
We build our model on the PyTorch with a NVIDIA 3090Ti GPU. The Adam Optimizer is utilized with a learning rate of 0.001. The TIU has three blocks. In each block, kernel-size is 3, dropout rate is 0.1 and dilation rate is $2^{i-1}$ in each layer $i$ ($i<4$).The linear has 256 units. For LTF, the hidden unit size is 256. The critic is a three-layer FCN with 256 units. In the DLN, a three-block TCN is used with 4 layers and 0.2 dropout rate in each block. The lengths of the observed sequence and the predicted sequence are set to 25 frames. Note that different datasets and different actions are trained independently in our method.
We evaluate our method by measuring the mean per joint position error (MPJPE) after alignment of the root joint. In our experiments, we consider two kinds of prediction: short-term prediction (less than $400$ $ms$) and long-term prediction ($400-1,000$ $ms$). 

\subsection{Comparison with Existing Methods}
\textbf{Results on H3.6M.} \quad
We benchmark our method against exist SOTA methods in Table \ref{tab:tab01} including res-GRU \cite{Res-gru}, HPGAN \cite{HPgan}, BiGAN \cite{Bigan}, HMR \cite{HMR}, LTD \cite{LTD}, DMGNN \cite{DMGNN}, HRI \cite{HRI}, MSR-GCN \cite{MSGNN}, SPGSN \cite{PGS} and FDU\cite{FDU}. 
HPGAN and BiGAN are classical GAN-based approaches that leverage the improved WGAN to predict both deterministic and non-deterministic human motions. However, the distribution exploration process inherent to these methods often impedes effective convergence, resulting in suboptimal accuracy. According to the quantitative results presented in Table \ref{tab:tab01}, our method enhances the performance of these networks by 37\% $\uparrow$, validating the efficacy of our proposed WGAN and brain-inspired learning.
Res-GRU and HMR are robust RNN-based methods. Despite their strengths, these methods are prone to error accumulation. In the direction action scenario, our method boosts short-term prediction accuracy by 100.6\% $\uparrow$ and long-term prediction accuracy by 95\% $\uparrow$. These improved results clearly demonstrate the viability of employing TRNs to address temporal challenges in motion prediction.
LTD, DMGNN, HRI, MSR-GCN, and FDU excel in terms of performance. LTD encodes temporal dependencies in trajectory space and utilizes a feed-forward network for HMP. DMGNN extracts features at individual scales and merges them using a multi-scale graph approach. HRI introduces motion attention mechanisms to extract similarities between the current motion context and historical motion sub-sequences, effectively avoiding pose similarity issues. MSR-GCN employs a multi-scale spatio-temporal graph to model motion relationships. These methods utilize various advanced strategies to tackle HMP, including trajectory space encoding, motion attention, and graphs. FDU achieves stable prediction results through a decomposition-aggregation two-stage strategy in frequency representation learning. Compared to these methods, our approach surpasses all others across all actions.
Besides quantitative evaluation, we conduct a visual comparison of the performance of SOTA methods. As illustrated in Fig.\ref{fig:3}, our method consistently maintains high fidelity to the ground truth in both short-term and long-term predictions. 

\begin{table}[t]
	\centering
\resizebox{.5\textwidth}{!}{
\renewcommand\scriptsize{7.0t}
 
		\begin{tabular}{cccccccccccccccccc}
			\hline
                Time (ms) & 80 & 160 & 320 & 400 & 1,000 & 80 & 160 & 320 & 400 & 1,000  \\ 
                \hline
			& \multicolumn{5}{c}{Bowling}  & \multicolumn{5}{c}{Golf} \\
			
			res-GRU   & 18.5& 33.9& 48.1& 49.0& 106.3 &12.9& 23.8& 40.2& 60.1& 77.5 &\\ 
			
			BiGAN  & 16.5& 30.5& 47.2& 48.8& 91.5 &8.7& 16.3& 30.1& 37.8& 76.6\\ 
			
			LTD  &14.3& 25.5& 48.4& 62.6& 109.0 &3.5& 6.7& 12.0& 15.8 &54.4 \\ 
			
			MSR-GCN   & 13.1& 22.1& 37.2& 55.8& 97.7 &3.4& 6.2& 11.2& 13.8& 47.3  \\ 
			
			Ours & \textbf{10.5} & \textbf{19.3}& \textbf{35.5} & \textbf{46.8} & \textbf{89.3} 
			& \textbf{2.5} & \textbf{6.0 }& \textbf{10.9 }& \textbf{12.8} & \textbf{44.5}\\ 
			\hline
		&\multicolumn{5}{c}{Tennis}  & \multicolumn{5}{c}{jump}\\
          res-GRU   &   17.4& 20.0& 27.3& 36.7& 50.2 & 20.3 & 39.5& 71.3& 84.0& 129.6\\ 
          BiGAN & 14.3& 16.3& 18.0& 20.2& 27.5 &12.1& 21.8& 41.9& 52.9& 94.6\\
            LTD &25.5& 36.7& 39.3& 39.9& 58.2 &11.3& 21.5& 44.2& 55.8 &117.5\\
        MSR-GCN  & 15.2& 19.7& 23.3& 35.8& 47.4 &10.3& 21.1& 42.7& 50.9& 91.4\\
        Ours & \textbf{11.1} & \textbf{14.3} & \textbf{17.2} & \textbf{18.8} & \textbf{25.4}
			& \textbf{7.9}& \textbf{15.1}& \textbf{30.5} & \textbf{41.2} & \textbf{85.3} \\
            \hline
	\end{tabular}}
\caption{Performance evaluation on G3D dataset.}
\label{tab:tab03}
\end{table}

\textbf{Results on CMU.} \quad
We further investigate our method on the CMU on 8 actions with results reported in Table \ref{tab:tab02}. 
Four methods with publicly available results are compared: res-GRU \cite{Res-gru}, BiGAN \cite{Bigan}, LTD \cite{LTD}, MSR-GCN \cite{MSGNN}. Quantitative analysis clearly indicates that our method outperforms all actions. These empirical findings reaffirm the superiority of our approach for HMP in both short-term and long-term scenarios. The consistent and significant performance improvements observed on the two benchmark datasets underscore the robustness of our method. Furthermore, as illustrated in Fig. \ref{fig:3}, our method produces significantly enhanced visual results on the CMU dataset, further demonstrating its efficacy.

\textbf{Results on G3D.}\quad
The results on G3D are reported in Table \ref{tab:tab03}. We also compare with the methods that are similar to CMU. From the quantitative evaluation, Our method achieves SOTA performance, demonstrating the effectiveness and robustness of our proposed method for both short-term and long-term predictions. 


\begin{table}[t]
	
	\centering
	\resizebox{.45\textwidth}{!}{
        \renewcommand\scriptsize{7.0t}
		\begin{tabular}{ccccccccccc}
			\hline
			HVM & HIM-TRN & HIM-DLN & 80 & 160 &320 & 400 & 1,000\\ \hline
			& $\checkmark$ & $\checkmark$
			& 16.8 & 28.1& 42.2 & 57.2& 121.9\\
            $\checkmark$& &$\checkmark$
			& 10.8 & 16.1 & 32.3 &47.1 & 101.0\\
            $\checkmark$&$\checkmark$ &
			& 8.1 & 15.9 & 30.0 & 45.1 & 98.7\\
			$\checkmark$&$\checkmark$ &$\checkmark$
			& \textbf{7.8} & \textbf{13.5} & \textbf{27.2} & \textbf{43.1} & \textbf{94.1}\\
			\hline
	\end{tabular}}
 \caption{Ablation studies on H3.6M dataset.}
 \label{tab:tab04}
\end{table}

\textbf{Ablation Experiments.}
We further study the influence of individual components in our framework through the following ablation studies. Experiments verify human-like vision module (HVM) and human-like inference module (HIM) on the H3.6M, as shown in Table \ref{tab:tab04}. 
First, we removed HVM and used the general GCN as the encoder. Without the HVM module, the performance of the model deteriorates dramatically. This clearly reflects that utilizing HVM to encode human motion significantly boosts accuracy for both short-term and long-term predictions. 
Next, to show the effectiveness of HIM, we replaced the TRN with GRU. Experimental results are reported in Table \ref{tab:tab04}.  Compared with the results of our method demonstrates the validity of the TCNs. 
Finally, we remove DLM directly. The DLM can effectively improve the predictive performance of the basic network.
The results of these ablation experiments show the contribution of each module that constitutes our method: 1) the HVM contributes to better encoding human motion and plays a crucial role in motion prediction. 2) The TRN captures temporal dependencies between joints, which is also important for generating accurate predictions. 3) The DLM contributes to enhancing trained model performance.

\section{Conclusion}  
In this paper, we propose a Human-like Vision and Inference System, which attempts to handle HMP issues in three aspects. 
1) Our approach is designed to simulate human observation and predict future motion with a human-like vision module and a human-like inference module. 
2) Simulating human visual perception, a human-like vision system is designed which can adequately capture spatio-temporal dependencies as well as global and local information.
3) A multi-step training strategy is proposed for simulating human-like inference, which simulates a spontaneous learning process to deal with the conventional prediction process and a deliberate learning process to improve the performance of difficult-to-train joints, respectively.  
Experimental results demonstrate that our approach significantly outperforms existing methods in both short-term and long-term HMP tasks.

\section{Acknowledgments}
This work is supported by the National Natural Science Foundation of China (No. 62372402), and the Key R\&D Program of Zhejiang Province (No. 2023C01217).

\bibliography{aaai25}

\end{document}